\documentclass[10pt, conference]{ieeeconf}      

\IEEEoverridecommandlockouts                              

\usepackage[nolist]{acronym}
\usepackage{graphicx}
\usepackage{caption}
\usepackage{subcaption}
\usepackage{bm}
\usepackage{amsfonts}
\usepackage{amsmath}
\usepackage{mathtools}
\usepackage{pdfcomment}
\usepackage{booktabs}
\usepackage{array}
\usepackage{makecell}
\usepackage{multirow}
\usepackage[table]{xcolor}
\usepackage{hhline}
\usepackage{hyperref}
\usepackage{tikz}

\DeclareMathOperator*{\argmax}{argmax}

\newcommand\copyrighttext{%
  \footnotesize \textcopyright 2021 IEEE. Personal use of this material is permitted.
  Permission from IEEE must be obtained for all other uses, in any current or future 
  media, including reprinting/republishing this material for advertising or promotional 
  purposes, creating new collective works, for resale or redistribution to servers or 
  lists, or reuse of any copyrighted component of this work in other works. DOI: 10.1109/ICRA48506.2021.9561206}
\newcommand\copyrightnotice{%
\begin{tikzpicture}[remember picture,overlay]
\node[anchor=south,yshift=10pt] at (current page.south) {\fbox{\parbox{\dimexpr\textwidth-\fboxsep-\fboxrule\relax}{\copyrighttext}}};
\end{tikzpicture}%
}

\title{\LARGE \bf
Robot Program Parameter Inference via Differentiable Shadow Program Inversion
}

\author{Benjamin Alt$^{1,2}$, Darko Katic$^{1}$, Rainer Jäkel$^{1}$, Asil Kaan Bozcuoglu$^{2}$ and Michael Beetz$^{2}$
\thanks{\raggedright$^{1}$ArtiMinds Robotics, Karlsruhe, Germany {\tt\footnotesize benjamin.alt@artiminds.com}}%
\thanks{$^{2}$Institute for Artificial Intelligence (IAI), University of Bremen, Germany}%
}

\let\oldmaketitle\maketitle
\renewcommand{\maketitle}{\oldmaketitle\setcounter{footnote}{2}}

\begin{document}

\maketitle
\copyrightnotice

\thispagestyle{empty}
\pagestyle{empty}

\acrodef{dp}[$\partial$P]{differentiable programming}
\acrodef{dsl}[DSL]{domain-specific language}
\acrodef{tcp}[TCP]{tool center point}
\acrodef{gru}[GRU]{Gated Recurrent Unit}
\acrodef{rnn}[RNN]{Recurrent Neural Network}
\acrodef{dag}[DAG]{directed acyclic graph}
\acrodef{nnii}[NNII]{Neural Network Iterative Inversion}
\acrodef{es}[ES]{Evolution Strategies}
\acrodef{pcb}[PCB]{printed circuit board}
\acrodef{tht}[THT]{through-hole technology}
\acrodef{vr}[VR]{virtual reality}
\acrodef{mp}[MP]{Movement Primitive}
\acrodef{sp}[SP]{Shadow Primitive}
\acrodef{artm}[ARTM]{ArtiMinds Robot Task Model}
\acrodef{rl}[RL]{reinforcement learning}
\acrodef{tamp}[TAMP]{Task and Motion Planning}
\acrodef{dmp}[DMP]{Dynamic Movement Primitive}
\acrodef{spi}[SPI]{Shadow Program Inversion}
\acrodef{fts}[FTS]{force-torque sensor}

\begin{abstract}
  Challenging manipulation tasks can be solved effectively by combining individual robot skills, which must be parameterized for the concrete physical environment and task at hand. This is time-consuming and difficult for human programmers, particularly for force-controlled skills. To this end, we present \ac{spi}, a novel approach to infer optimal skill parameters directly from data. \ac{spi} leverages unsupervised learning to train an auxiliary differentiable program representation (``shadow program'') and realizes parameter inference via gradient-based model inversion. Our method enables the use of efficient first-order optimizers to infer optimal parameters for originally non-differentiable skills, including many skill variants currently used in production. \ac{spi} zero-shot generalizes across task objectives, meaning that shadow programs do not need to be retrained to infer parameters for different task variants. We evaluate our methods on three different robots and skill frameworks in industrial and household scenarios. Code and examples are available at \url{https://innolab.artiminds.com/icra2021}.
\end{abstract}

\section{Introduction}
Combining individual robot skills to solve complex tasks has established itself as one of the primary programming paradigms in robotics. A large variety of skill variants such as \acp{dmp} \cite{schaal_dynamic_2006}, \ac{tamp} operators \cite{kaelbling_hierarchical_2011} or generalized manipulation strategies \cite{jakel_learning_2013} have been proposed, all of which allow the behavior of skills to be adapted to the task at hand by a set of skill parameters such as velocities, scaling factors or via-points. Finding appropriate values for these parameters is difficult and typically requires manual tweaking. Consider a force-controlled spiral search skill, where the robot executes a spiral motion until a drop in forces at the \ac{tcp} indicate that a hole has been found. The optimal parameters, i.e. spiral orientation and extents, velocity, acceleration and pushing force, to maximize the likelihood of finding a hole without sacrificing too much time, depend on the probability distribution of the hole position and the physical properties of the surfaces. For a human programmer, tuning these parameters involves much trial and error.\\
\Ac{dp} provides an elegant solution: If a program is fully differentiable, it allows for the gradient-based optimization of program parameters with respect to nearly arbitrary objective functions over the program's outputs \cite{baydin_automatic_2018, innes_differentiable_2019}. However, most skill libraries used in practice are not differentiable. Particularly force-controlled skills such as spiral search or moment-free insertion are often implemented using highly performant but non-differentiable force controllers provided by robot manufacturers.\\
\begin{figure}
  \centering
  \includegraphics[width=\linewidth]{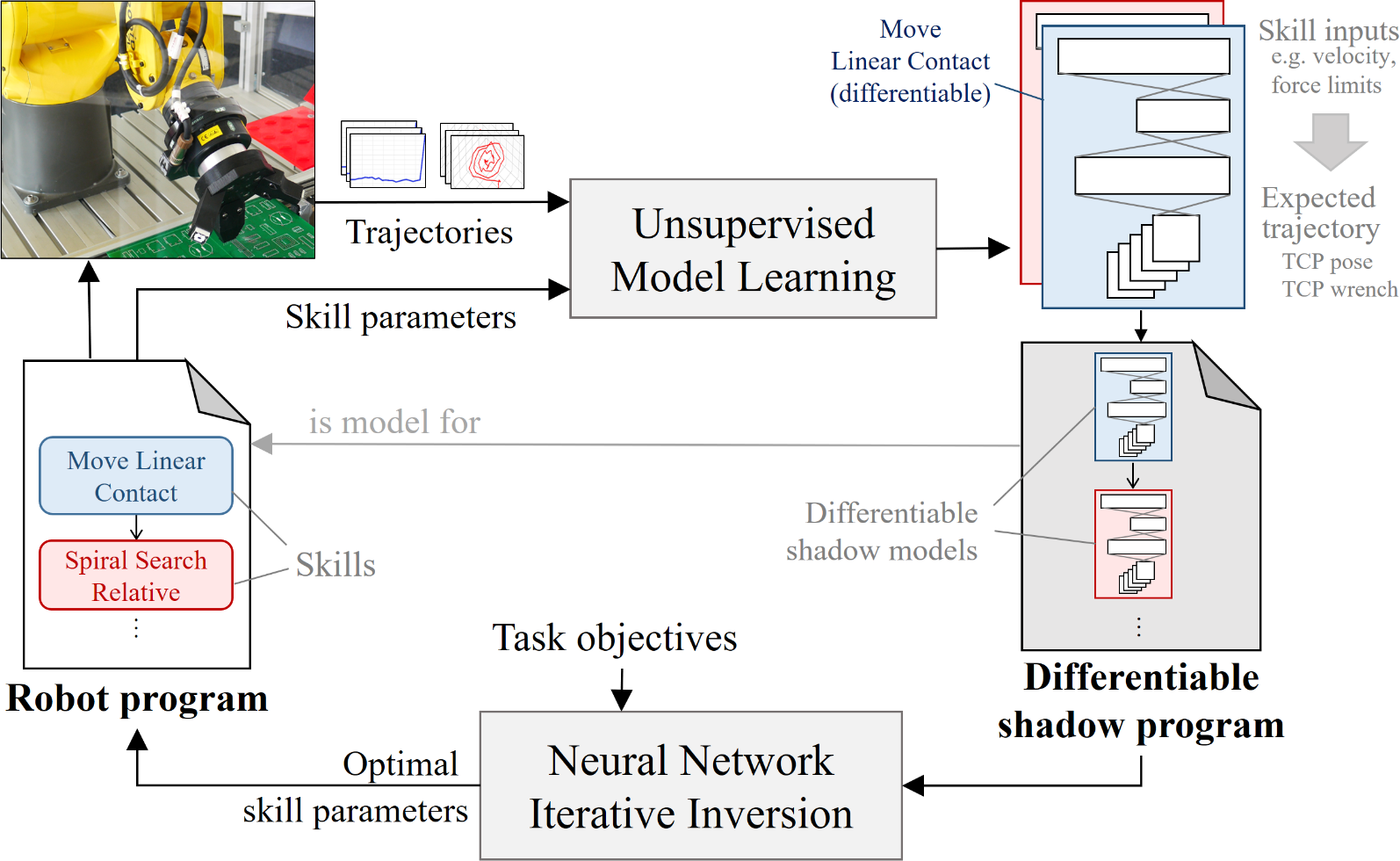}
  \caption{Parameter inference via \acl{spi}. By inverting learned differentiable shadow models (right) of a sequence of robot skills (left), optimal program parameters can be inferred directly from data.}
  \label{fig:system}
\end{figure}
In this paper, we present \acf{spi}, a novel \ac{dp}-based method of inferring optimal parameters for robot programs of non-differentiable skills. We propose to learn a differentiable surrogate (called \textit{shadow model}) of each skill, which is trained via unsupervised learning to predict the expected trajectory (TCP poses and wrenches) when executing the skill for a given set of parameters. Just like skills can be chained to solve complex tasks, their shadow models can be chained to form a differentiable \textit{shadow program}. We then apply a gradient-based neural network inversion technique to the shadow program to jointly infer the skill parameters which maximize a set of task objectives. By maintaining a one-to-one relationship between the original robot skills and their differentiable shadows, the optimized program parameters can be transferred back to the original skills, which are used for execution on the robot.\\
By conducting gradient-based parameter inference over differentiable surrogates rather than the actual skills, our method can be applied optimize parameters for widely used skill representations such as \acp{dmp}, which do not have to be differentiable. Moreover, it generalizes in a zero-shot manner across task objectives, avoiding retraining when task objectives change. We demonstrate the broad applicability of our approach on three use cases from industrial and household robotics, involving diverse skill representations ranging from high-level generalized manipulation strategies to low-level \acp{dmp} and the URScript robot API \cite{noauthor_urscript_2018}. To our knowledge, \ac{spi} is the first application of gradient-based model inversion to robot skill parameterization.\\
Our main contributions can be summarized as follows:
\begin{enumerate}
  \item \textbf{Shadow models}, differentiable surrogate representations of possibly non-differentiable skills.
  \item \textbf{Shadow programs}, end-to-end differentiable representations of complete skill-based robot programs.
  \item \textbf{\acl{spi}}, an algorithm to efficiently infer optimal skill parameters via gradient-based model inversion.
  \item \textbf{Evaluation} of \acl{spi} on three different robots and real-world use cases.
\end{enumerate}

\section{Related Work}
\subsubsection{Robot skill parameter inference}
Several approaches for the automatic optimization of robot skill parameters have been proposed, which rely on gradient-free optimization techniques such as evolutionary algorithms \cite{liang_optimization_2017, urieli_optimizing_2011, marvel_automated_2009} or Bayesian optimization \cite{berkenkamp_bayesian_2020, calandra_bayesian_2016, akrour_local_2017} due to the non-differentiability of most skill libraries and frameworks. Gradient-free approaches require frequent execution of the skills during optimization, which is a time-consuming process if done on real robot systems, has to be repeated whenever the task objectives change and often require good initial parameterizations. We propose to use a gradient-based optimizer on a differentiable surrogate model of the skills, which avoids these issues.

\subsubsection{Unsupervised representation learning}
Recent robot learning approaches such as Visual Foresight \cite{finn_deep_2017}, Adversarial Skill Networks \cite{mees_adversarial_2020}, Time Reversal \cite{nair_time_2020} or ``Learning from Play'' \cite{lynch_learning_2019} propose self-supervised or unsupervised learning of a \textit{predictive model}, i.e. a model of skill inputs to a latent trajectory representation, from which a policy to solve the task is then derived. We take a similar approach to parameterize skill-based robot programs by learning a differentiable skill model and exploit its differentiability to infer optimal program parameters. Like most approaches relying on self-supervised representation learning, ours generalizes across task variants without requiring additional training.

\subsubsection{Differentiable programming}
\Acf{dp} proposes to express programs as differentiable computational graphs, which permit the optimization of program parameters via reverse-mode automatic differentiation and gradient-based optimization \cite{innes_differentiable_2019, baydin_automatic_2018}. From a \ac{dp} perspective, neural networks can be considered a type of differentiable program, and can be combined to computational graphs-of-networks \cite{parisotto_neuro-symbolic_2016, reed_neural_2015, rabinovich_abstract_2017} as well as hybrid architectures combining neural networks with differentiable hand-written algorithms or data structures \cite{graves_hybrid_2016, gaunt_differentiable_2017, bosnjak_programming_2017, kurach_neural_2015, garnelo_conditional_2018}. In the domain of robotics, \ac{dp} has been realized in the form of hybrid skill representations such as Conditional Neural Movement Primitives \cite{seker_conditional_2019, akbulut_acnmp_2020},  Deep Movement Primitives \cite{pervez_learning_2017} or Differentiable Algorithm Networks \cite{karkus_differentiable_2019}, which combine differentiable algorithmic priors with neural networks and can be combined to complex robot programs in a modular fashion. Like most prior work on \ac{dp} in robotics, we provide modular interfaces to combine differentiable functional blocks to complex programs, and combine hand-written computational graphs with neural networks to simplify the learning problems. In contrast to prior work, however, we rely exclusively on unsupervised learning, which greatly simplifies data collection and model training. In work similar to ours, Zhou et al. \cite{zhou_movement_2020} propose to generate \ac{dmp} parameters directly with a Mixture Density Network. We instead propose to learn a differentiable model of the \ac{dmp} and to infer the optimal skill parameters via model inversion after training. The advantage is that no retraining is required when the task objective changes. Moreover, our approach can be applied to near-arbitrary skill representations beyond \acp{dmp} and allows the optimization of parameters for complex programs composed of multiple skills.

\section{Definitions \& Problem Statement}

\subsection{Definition: Skill-based Robot Program}
\label{sec:skill_based_robot_program_definition}
We define a skill-based robot program as a \ac{dag} of skills, where each skill is defined as a function $f: \mathcal{X} \times \mathcal{S} \to \mathcal{S} \times \mathcal{Y}$ with the space of the skill's exposed input parameters $\mathcal{X}$, $\mathcal{S}$ the state space comprising the current poses of coordinate frames relevant to the task and $\mathcal{Y}$ the trajectory space comprising \ac{tcp} poses and wrenches. This definition allows to treat a skill as a black box which maps a vector of inputs $\bm{x} \in \mathcal{X}$ and prior state $\bm{s}_{in} \in \mathcal{S}$ to posterior state $\bm{s}_{out} \in \mathcal{S}$ and trajectory $\bm{Y} \in \mathcal{Y}$. It covers \acp{dmp} \cite{schaal_dynamic_2006}, generalized manipulation strategies \cite{jakel_learning_2013} or any skill variant which takes some inputs $\bm{x}$ and moves the robot. This definition allows to design a universal differentiable shadow model architecture capable of representing them.

\subsection{Problem Statement: Robot Program Parameter Inference}
Sequentially chained skills form complex robot programs by propagating the posterior state $\bm{s}_{i, out}$ of the $i$th skill in the sequence to the prior state $\bm{s}_{i+1, in}$ of the subsequent skill. We postulate the Markov property, i.e. all relevant context information is captured in the start state $\bm{s}_{in}$. Such a skill-based robot program realizes the function $P: \mathcal{S} \times \mathcal{X}^n \to \mathcal{S} \times \mathcal{Y}$. We seek to optimize a task-dependent objective function $\Phi: \mathcal{Y} \to \mathbb{R}$, which assigns a real-valued score to a trajectory. For a program containing a spiral search skill, $\Phi$ might assign high scores to trajectories which exhibit the characteristic drop in forces indicating that the hole has been found. For a given program $P$ and initial state $\bm{s}_{in}$, we seek to solve the inverse problem $\bm{x^*} = \argmax_{\bm{x} \in \mathcal{X}^n} \Phi(P(\bm{s}_{in}, \bm{x}))$, i.e. to find skill parameters which maximize $\Phi$. In the spiral search example, $\bm{x^*}$ contains the velocities, accelerations, spiral extents and force setpoints which maximize the likelihood of the hole being found. Due to the high dimensionality of the combined input space $\mathcal{X}^n$ of complex programs, learning to directly compute $\bm{x^*}$ would require prohibitive amounts of supervised training data. We propose to instead train a differentiable model $\hat{P}$ (called \textit{shadow program}) of $P$, and to iteratively approximate $\bm{x^*}$ by gradient descent over $\hat{P}$.

\section{Parameter Inference via Shadow Program Inversion}
\label{sec:sps}

We propose a three-step process to infer skill parameters for a skill-based robot program: (1) Construction of a differentiable surrogate (shadow program, $\hat{P}$), which predicts the expected trajectory (\ac{tcp} poses and wrenches) given input parameters for all skills in the program; (2) unsupervised training of the learnable components of this surrogate; and (3) inference of optimal skill parameters via gradient-based model inversion (cf. figure \ref{fig:system}).

\subsection{Shadow Model \& Shadow Program Architecture}
\begin{figure}
  \centering
    \includegraphics[width=\linewidth]{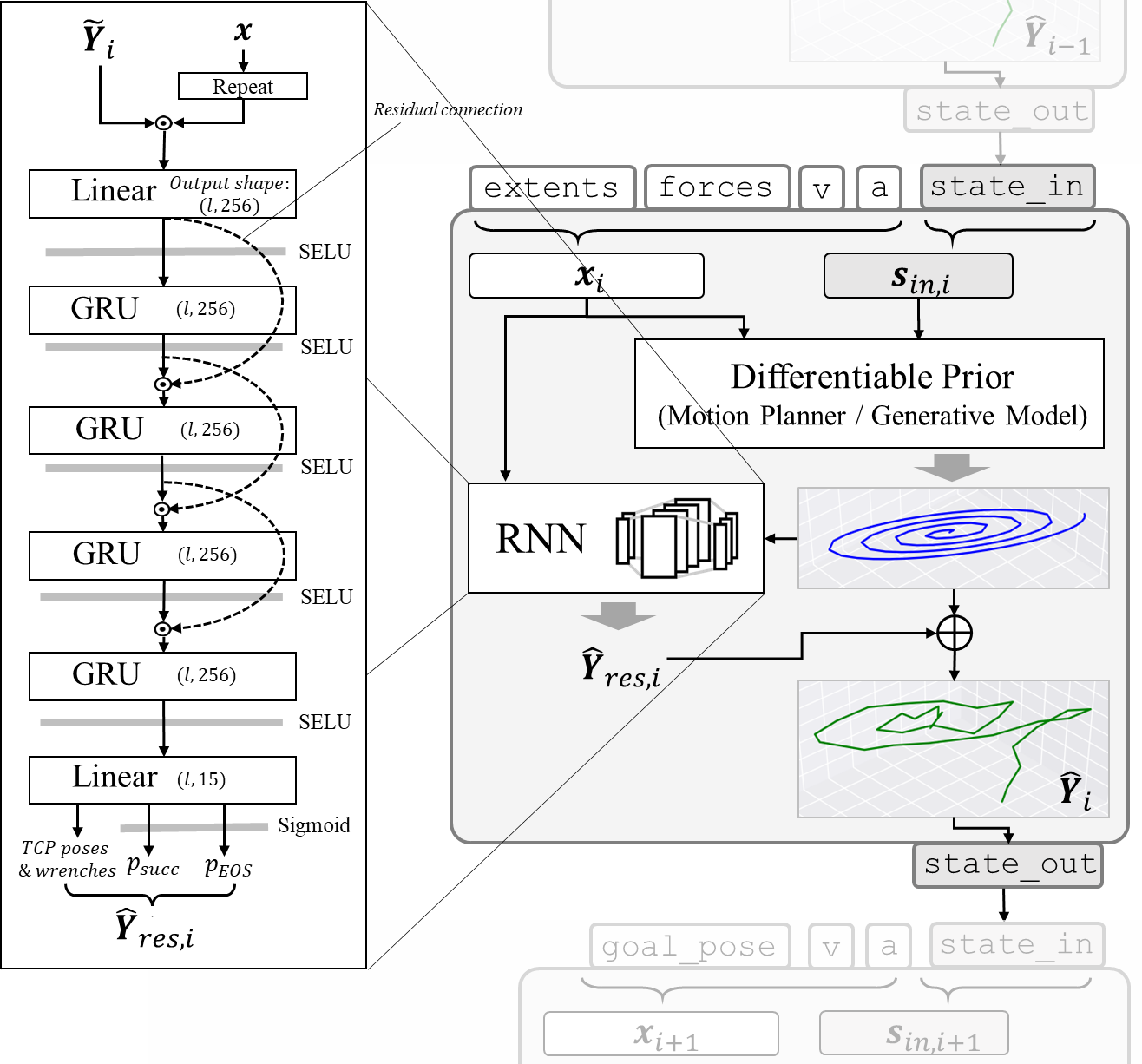}
    \caption{A shadow model of a spiral search skill as part of a larger shadow program.}
    \label{fig:rnn}
\end{figure}

For a given skill-based robot program (such as a sequence of \acp{dmp}) $P$, whose parameters are to be inferred, we begin by constructing the differentiable shadow program $\hat{P}$. To that end, we instantiate a differentiable \textit{shadow model} for each skill in $P$. Because we only consider skills which meet definition \ref{sec:skill_based_robot_program_definition}, we can propose one single, differentiable shadow model architecture which can model any skill. This architecture is illustrated in figure \ref{fig:rnn}.\\
A shadow model of the $i$-th skill in a skill-based robot program is a differentiable computational graph which computes the expected trajectory $\bm{\hat{Y}}_i$ for a given input vector $\bm{x}_i$ and start state $\bm{s}_{in, i}$. We represent $\bm{\hat{Y}}_i$ as a sequence of samples $(\bm{\hat{Y}}_i)_t$, $ 0 \leq t < |\bm{\hat{Y}}_i|$, each of which contains the current success probability $p_{succ} \in [0, 1]$, the probability $p_{EOS} \in [0, 1]$ of the action being completed at time $t$, as well as the \ac{tcp} pose and wrench at time $t$. In figure \ref{fig:rnn}, $\bm{x}_i$ contains the input parameters of a force-controlled spiral search skill, and $\bm{s}_{in, i}$ is set to the final state $\bm{s}_{out, i-1}$ of the previous skill. The inclusion of $p_{succ}$ in $\mathcal{Y}$ permits the inference of skill parameters with respect to skill-specific success metrics. For a spiral search, it allows to specify an objective function to maximize the probability of finding the hole (cf. equation \ref{eq:g_fail} and experiment \ref{sec:experiment_3}).\\
Echoing work in \ac{dp} integrating algorithmic priors and deep learning \cite{bhardwaj_differentiable_2020, garnelo_conditional_2018}, we do not predict the expected trajectory $\bm{\hat{Y}}_i$ end-to-end. Instead, we bootstrap an initial prior trajectory estimate $\bm{\widetilde{Y}}_i$ using a differentiable motion planner,\footnote{For our experiments (see section \ref{sec:experiments}), we implemented simple differentiable planners for linear motions, spiral motions and gripper motions by reimplementing parts of orocos-kdl \cite{noauthor_orocos_2020} and urdfpy \cite{matl_urdfpy_2020} with PyTorch \cite{paszke_pytorch_2019}.} which produces a crude approximation of the trajectory without interactions with the environment such as moving objects or applying contact forces. We found that explicitly incorporating prior knowledge greatly reduced the amount of training data required, particularly for long trajectories. For skills for which no such algorithmic prior exists, we instead use a generative neural network to bootstrap a prior trajectory from $\bm{x}_i$ and $\bm{s}_{in, i}$ \cite{graves_generating_2014}. The posterior trajectory $\bm{\hat{Y}}$ is the sum of the prior and the output of a deep residual \ac{gru} \cite{cho_learning_2014}, which predicts the residual trajectory $\bm{\hat{Y}}_{res, i}$ containing the context-dependent information about interactions with the environment, such as (expected) forces and torques. For the spiral search skill, for example, the residual \ac{gru} learns to predict when and where a hole is likely to be found, and how the expected trajectory then deviates from the prior $\bm{\widetilde{Y}}_i$.\\
The shadow model architecture is sufficiently flexible to model simple and complex skills with varying numbers and types of parameters. Note that the ``signature'' (layout of the parameter vector $\bm{x}$) of a shadow model exactly matches that of the skill it models. This permits the transfer of the inferred parameters back to the original skill after inference (cf. \ref{sec:shadow_program_inversion}). Aside from the optional differentiable motion planner, which will differ from skill to skill, the proposed shadow model architecture can model any skill which meets the skill definition in section \ref{sec:skill_based_robot_program_definition}. This allows the automatic construction of a shadow program for any complex skill-based robot program by instantiating a shadow model for each skill, and connecting the posterior state $\bm{s}_{out}$ of each shadow model to the prior state $\bm{s}_{in}$ of the next.

\subsection{Unsupervised Shadow Model Learning}
Because shadow models are forward models of semi-symbolic skills, they can be trained end-to-end on tuples $(\bm{x}, \bm{s}_{in}, \bm{Y})$\footnote{In our implementation, $\bm{s}_{out}$ can be deterministically computed from the output trajectory $\bm{\hat{Y}}$ and does not need to be learned explicitly.} to minimize the mean prediction error
\begin{gather}
\begin{aligned}
  \mathcal{L}_{pred}(\bm{\hat{Y}}, \bm{Y}) = &\:\frac{1}{\lvert \bm{\hat{Y}} \rvert} \sum_{n=1}^{\mathclap{\lvert \bm{\hat{Y}} \rvert = \lvert \bm{Y} \rvert}}\big(w_{pos} \mathcal{L}_{L2}((\bm{\hat{Y}})_{n,pos}, (\bm{Y})_{n,pos}) \\&\: + w_{ori} \mathcal{L}_{ori}((\bm{\hat{Y}})_{n,ori}, (\bm{Y})_{n, ori}) \\&\: + w_{ft} \mathcal{L}_{L2}((\bm{\hat{Y}})_{n,ft}, (\bm{Y})_{n,ft}) \\&\: + w_{succ} \mathcal{L}_{BCE}((\bm{\hat{Y}})_{n, p_{succ}}, (\bm{Y})_{n, p_{succ}}) \\&\: + w_{EOS}\mathcal{L}_{BCE}((\bm{\hat{Y}})_{n, p_{EOS}}, (\bm{Y})_{n, p_{EOS}})\big)\text{,}
\end{aligned}\raisetag{3\baselineskip}
\end{gather}
a weighted sum of the squared pointwise L2 distance $\mathcal{L}_{L2}$ for \ac{tcp} position and wrench, the pointwise binary crossentropy loss $\mathcal{L}_{BCE}$ for $p_{succ}$ and $p_{EOS}$, and the pointwise angle between quaternion-encoded \ac{tcp} orientations
\begin{equation}
\mathcal{L}_{ori}(\bm{\hat{y}}, \bm{y}) = \cos^{-1}(2\langle \bm{\hat{y}}_{ori}, \bm{y}_{ori} \rangle^2 - 1).\label{eq:l_ori}
\end{equation}
$\langle \bm{q}_1, \bm{q}_2 \rangle$ denotes the inner product of quaternions $\bm{q}_1$ and $\bm{q}_2$. $(\bm{Y})_{n, pos}$ denotes the position component of the $n$-th point on trajectory $\bm{Y}$.
Training data can be collected autonomously by sampling inputs $\bm{x}$ and initial states $\bm{s}_{in}$, executing the original (non-differentiable) skills and recording the resulting trajectories. In real-world settings, in which programs are executed repeatedly over long periods of time, this permits the efficient use of readily available unsupervised data and facilitates automatic finetuning of the model as new observations become available. Because of the Markov property, shadow models can be trained separately from one another, conditional on the start state $\bm{s}_{in}$, to form a library of trained, differentiable shadow models.

\subsection{Gradient-based Shadow Program Inversion}
\label{sec:shadow_program_inversion}
\begin{figure}
  \centering
  \includegraphics[width=\linewidth]{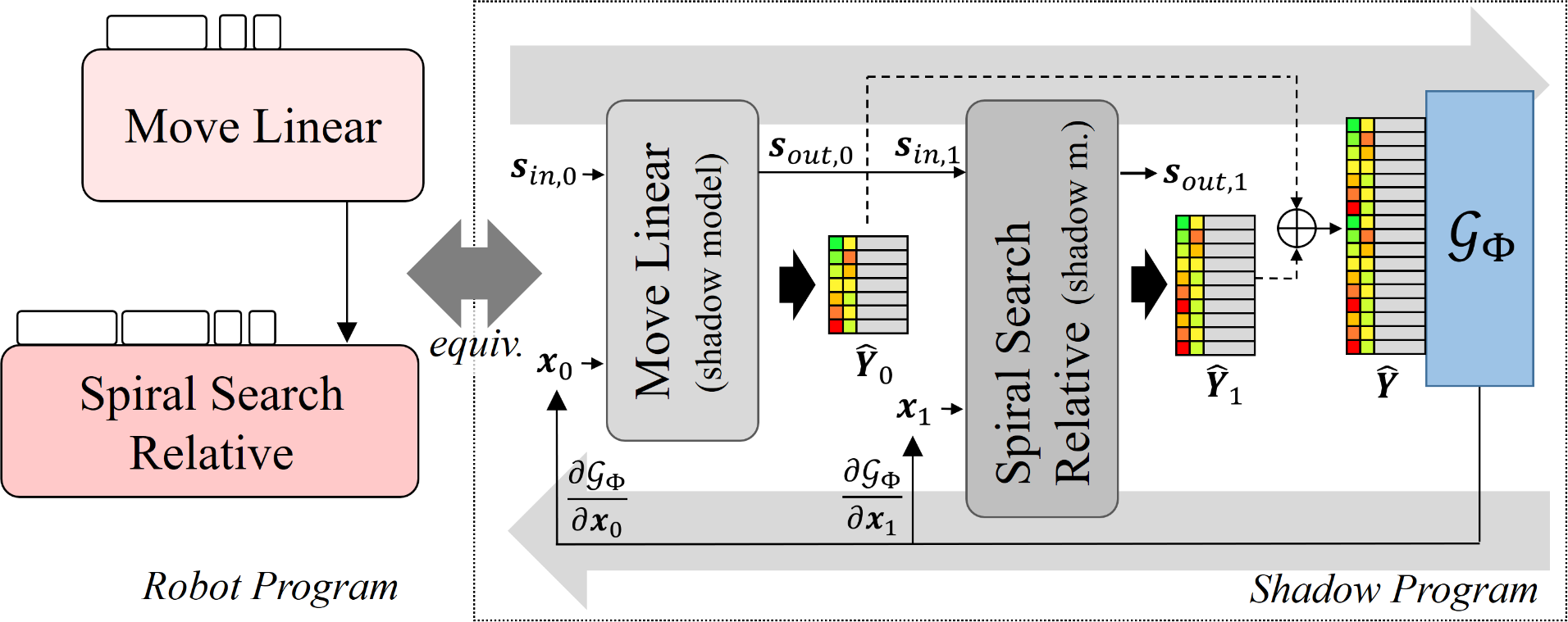}
  \caption{\ac{spi} (right) permits inference of skill parameters ($\bm{x}_0$ and $\bm{x}_1$) w.r.t. task objectives $\mathcal{G}_\Phi$ for non-differentiable robot programs (left).}
  \label{fig:neural_program}
\end{figure}

To infer optimal parameters for a given skill-based robot program, we construct the corresponding shadow program by sequentially chaining trained shadow models via $\bm{s}_{in}$ and $\bm{s}_{out}$ (see figure \ref{fig:neural_program}). The shadow program is a differentiable graph-of-graphs; its differentiability permits the efficient, gradient-based optimization of the input parameters $\bm{x}_i, 0 \le i < n$ of all $n$ skills of in the program with respect to an arbitrary, differentiable objective function $\mathcal{G}_\Phi$ for task objectives $\Phi$. To perform this optimization, we use \ac{nnii} \cite{hoskins_iterative_1992}: We randomly initialize the $\bm{x}_i$, perform a forward pass and compute the loss $\mathcal{G}_\Phi(\bm{\hat{Y}})$. We exploit the differentiability of the program graph to backpropagate the gradients $\frac{\partial \mathcal{G}_\Phi(\bm{\hat{Y}})}{\partial \bm{x}_i}$ and update the $\bm{x}_i$ according to the Adam update rule \cite{kingma_adam_2017}. Iterating until convergence yields the $\bm{x}_i$ which minimize $\mathcal{G}_\Phi$.\\
With differentiability w.r.t. $\bm{\hat{Y}}$ the sole requirement, a wide range of objective functions can be applied. In an industrial context, typical optimization targets include process metrics such as cycle time ($\mathcal{G}_{cycle}$), failure rate ($\mathcal{G}_{fail}$) or path length ($\mathcal{G}_{path}$). The inclusion of meta information $p_{EOS}$ and $p_{succ}$ in $\bm{\hat{Y}}$ permits the succinct expression of the corresponding objective functions:
\begin{equation}
  \mathcal{G}_{fail}(\bm{\hat{Y}})= {-} {\max}\Big(0, {\min}\big(\frac{1}{\lvert \bm{\hat{Y}} \rvert}\sum_{n=1}^{\lvert \bm{\hat{Y}} \rvert} (\bm{\hat{Y}})_{n, p_{succ}}, 1\big)\Big)\label{eq:g_fail}
\end{equation}
\begin{equation}
  \mathcal{G}_{cycle}(\bm{\hat{Y}})=\sum_{n=1}^{\lvert \bm{\hat{Y}} \rvert} \Big(1 - \sigma \big((\bm{\hat{Y}})_{n, p_{EOS}} - 0.5\big) * T \Big)\label{eq:g_cycle}
\end{equation}
\begin{gather}
  \begin{aligned}
\mathcal{G}_{path}(\bm{\hat{Y}}) =&\:\frac{1}{\lvert \bm{\hat{Y}} \rvert} \sum_{n=1}^{\lvert \bm{\hat{Y}} \rvert - 1} \Big( \lVert (\bm{\hat{Y}})_{n + 1, pos} - (\bm{\hat{Y}})_{n, pos} \rVert \\ +&\: \mathcal{L}_{ori}((\bm{\hat{Y}})_{n + 1, ori}, (\bm{\hat{Y}})_{n, ori})\Big)
  \end{aligned}
\end{gather}
$\sigma$ is the sigmoid function, $T$ a constant (here $T=100$) and $\mathcal{L}_{ori}$ as defined in eqn. \ref{eq:l_ori}. Joint optimization of multiple metrics at the same time can be realized by linear combination.\\
\ac{spi} as described above has several properties which make it both theoretically attractive and practically applicable in real-world scenarios:
\subsubsection{Asymptotic optimality} If $\mathcal{G}_\Phi$ is the objective function of the equivalent minimization problem to $\argmax_{\bm{x}} \Phi(P(\bm{s}_{in}, \bm{x}))$, $\bm{x}$ will approximate the optimal parameters $\bm{x^*}$, provided $\mathcal{G}_\Phi$ is convex, $P$ is faithfully approximated by the shadow program, gradients are bounded and the learning rate is small \cite{kingma_adam_2017}. In practice, near-optimal solutions can be reached in a few hundred iterations.
\subsubsection{Separation of learning from inference} Because the learning problem is reduced to learning a forward model of the program, parameter inference is decoupled from training and therefore very fast. Individual shadow models can be trained offline and combined to arbitrary shadow programs at inference time. Parameter inference itself does not require additional training, exploration or expensive policy search.
\subsubsection{Zero-shot generalization} By extension, our approach permits parameter inference with respect to arbitrary objective functions without requiring additional training examples. The same robot program can be optimized for different task objectives $\Phi$ by simply changing the loss function $\mathcal{G}_\Phi$ accordingly and rerunning \ac{nnii}.

\section{Experiments}
\label{sec:experiments}
To evaluate our approach in a wide variety of real-world applications, we apply \ac{spi} to infer program parameters from human demonstrations for pick-and-place tasks in a household scenario, impact force optimization for contact motions and the inference of spiral search heuristics in the context of electronics assembly.

\subsection{Parameter Inference for Complex Task Objectives}
\label{sec:experiment_1}
\begin{figure}
  \centering
  \begin{subfigure}[b]{.31\linewidth}
  \includegraphics[width=\linewidth]{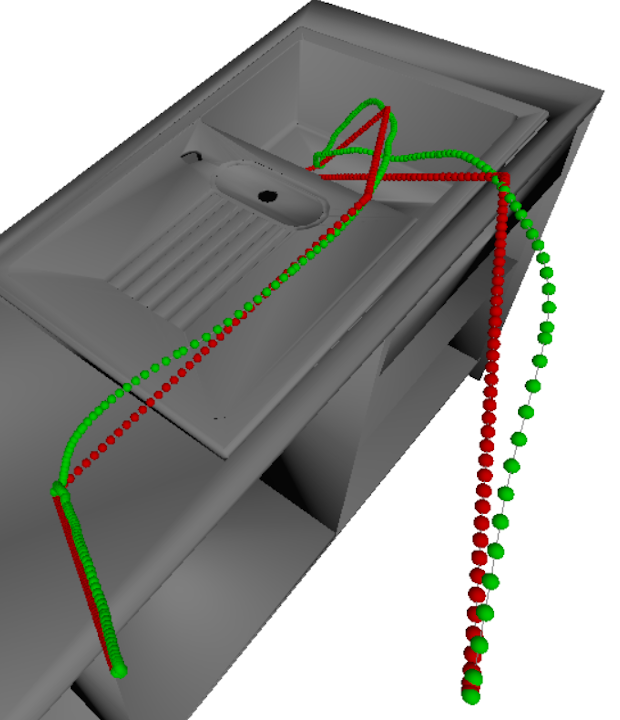}
  \end{subfigure}
  \begin{subfigure}[b]{.31\linewidth}
  \includegraphics[width=\linewidth]{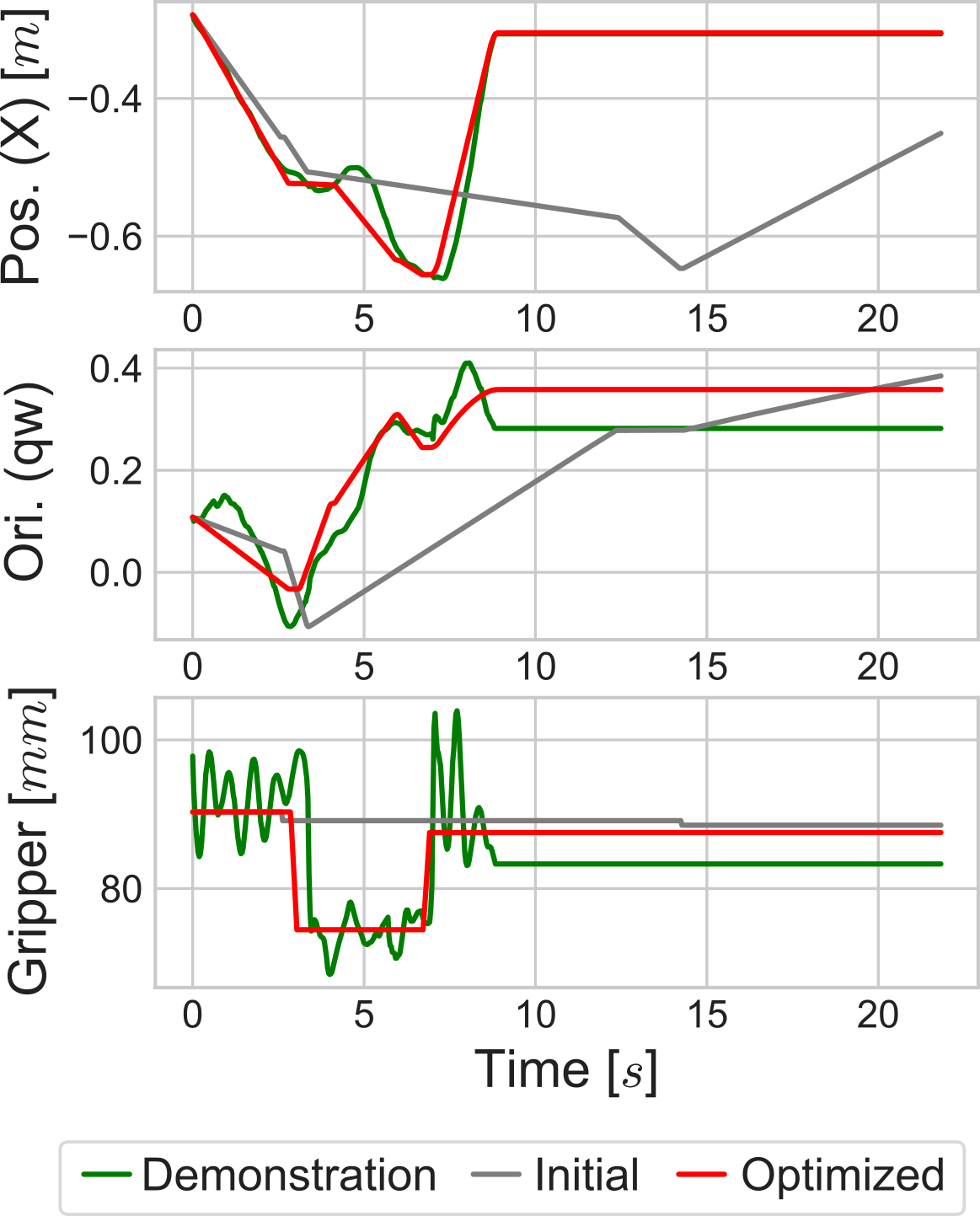}
  \end{subfigure}
  \begin{subfigure}[b]{.32\linewidth}
    \includegraphics[width=\linewidth]{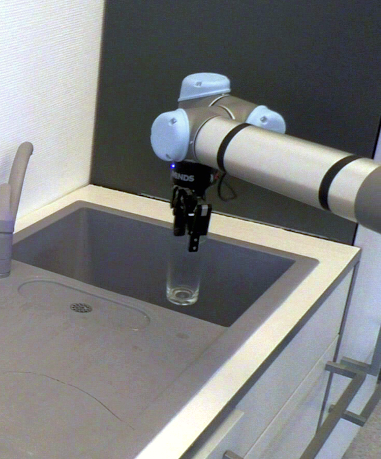}
  \end{subfigure} 
  \caption{Experiment \ref{sec:experiment_1}: Human demonstration (green) and trajectory after parameter inference (red) in 3D (left); 2D projections onto a position and orientation dimension and the gripper opening (middle); real-world execution (right).}
  \label{fig:vr_human_demonstration}
\end{figure}
In this experiment, we demonstrate the capacity of \ac{spi} to infer parameters with respect to complex task objectives from scratch. To that end, a household task of picking up a glass and depositing it in a sink is considered. Given only a program structure (a sequence of unparameterized skills), a set of parameters is to be inferred which closely approximates a human demonstration of the task. The program consists of a linear approach motion, opening the gripper, a sequence of 3 linear transfer motions, a skill to close the gripper and a linear depart motion.\footnote{The skill representation for which parameters are inferred in experiments \ref{sec:experiment_1}, \ref{sec:experiment_2a} and \ref{sec:experiment_3} is the \ac{artm} \cite{schmidt-rohr_artiminds_2013}, a non-differentiable industrial implementation of generalized manipulation strategies \cite{jakel_learning_2013}.} Skill parameters are initialized randomly and comprise the goal poses, velocities and accelerations of the linear motions as well as the target joint state and velocity of the gripper skills. Program parameters were inferred to minimize a combination of pointwise distance $\mathcal{G}_d$ between the \ac{tcp} and the demonstration as well as the demonstrated and predicted hand openings, and a grasp penalty $\mathcal{G}_g$ enforcing additional precision during grasps. Minimizing $\mathcal{G}_d$ and  $\mathcal{G}_g$ is a challenging optimization problem because the dynamics of the demonstration and predicted trajectories are vastly different, and \ac{spi} must implicitly adapt the velocities and accelerations of the skills first in order to make the predicted trajectories comparable to the demonstration. Four human demonstrations were collected in \ac{vr} using the KnowRob framework \cite{beetz_know_2018}. For this use case, the gripper state was included in $\mathcal{S}$ and $\mathcal{Y}$. Real-world experiments were conducted using a Universal Robots UR5 industrial manipulator and a Robotiq 2FG-85 parallel gripper. Results are shown in figure \ref{fig:vr_human_demonstration}. For each of the four demonstrations, parameter inference results in a robot motion which closely approximates the human demonstrations, but obeys the constraints such as linearity imposed by the skills. The results testify to the capacity of \ac{spi} to jointly infer skill inputs for realistic robot programs, even if the initial program parameters are far from the optimum.

\begin{figure}
  \centering
  \begin{minipage}{.49\linewidth}
    \includegraphics[width=\linewidth]{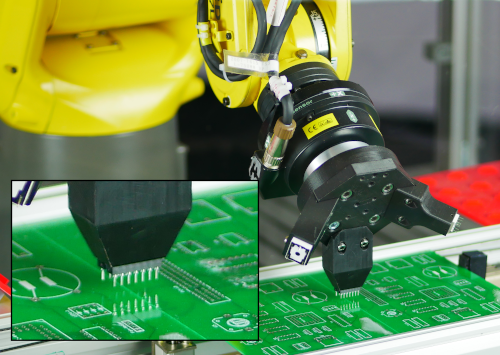}
    \vskip\baselineskip\vspace{-6pt}%
    \renewcommand{\tabcolsep}{2.6pt}
    \resizebox{\linewidth}{!}{
    \begin{tabular}{lccccccccc}
      \toprule
      $F_{goal}$ & $3N$ & $4N$ & $5N$ & $6N$ & $7N$ \\\midrule
      ARTM &  \makecell{0.69\\-75\%} & \makecell{0.54\\-69\%} & \makecell{0.54\\51\%} & \makecell{0.38\\-64\%} & \makecell{0.77\\-51\%}\\
      \bottomrule
     \end{tabular}}  
  \end{minipage}%
  \hfill
  \begin{minipage}{.49\linewidth}
    \includegraphics[width=\linewidth]{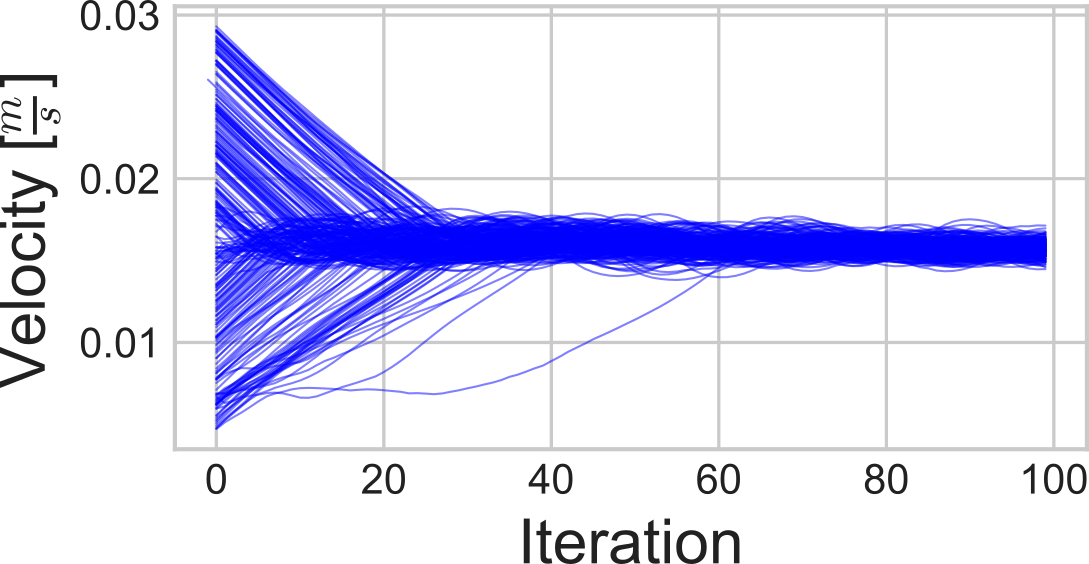}
     \vskip\baselineskip\vspace{-8pt}%
    \includegraphics[width=\linewidth]{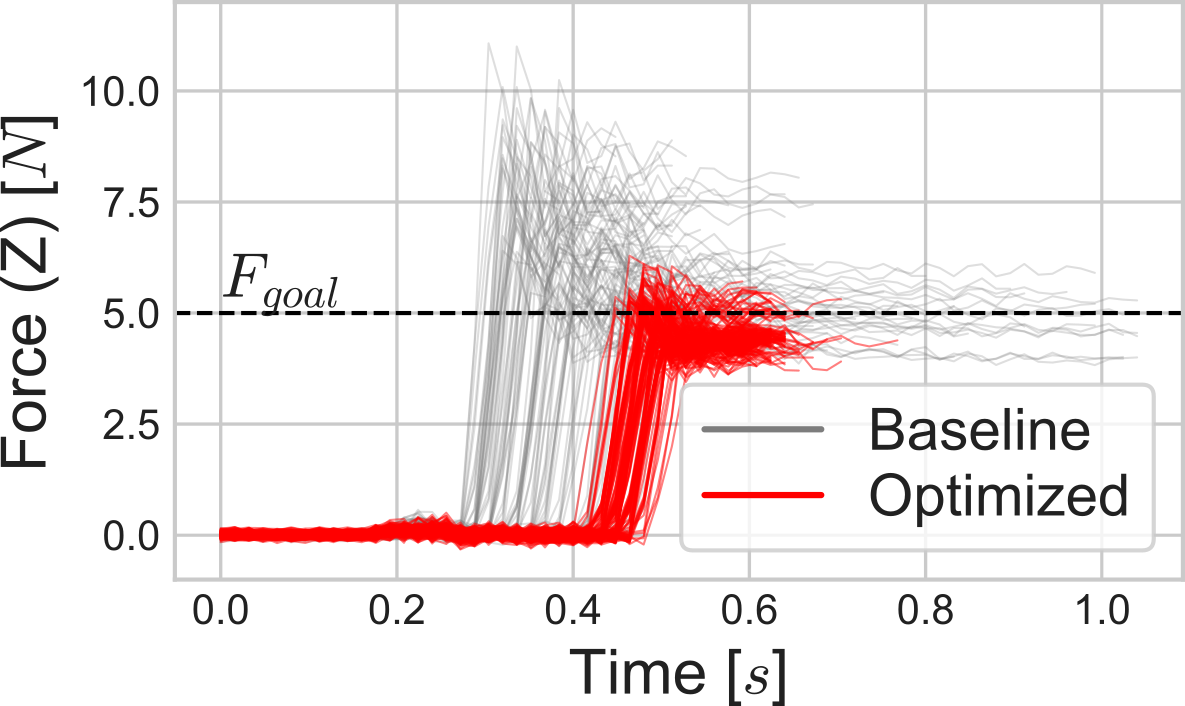}
    \end{minipage}
  \caption{Experiment \ref{sec:experiment_2}: Optimization of contact motions (bottom left). Each cell shows the mean deviation from $F_{goal}$ over 250 optimizations and the improvement over the initial parameterization. Right: Convergence behavior of \ac{spi} for the velocity parameter and resulting force trajectories for $F_{goal}$ = 5 $N$ for a linear motion ARTM skill.}
  \label{fig:results_mlrc}
\end{figure}
\begin{figure}
  \centering
  \includegraphics[width=\linewidth]{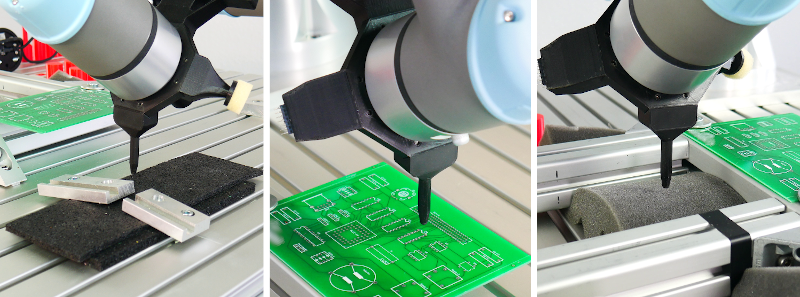}    
  \vskip\baselineskip\vspace{-8pt}%
  \renewcommand{\tabcolsep}{2.6pt}
  \resizebox{\linewidth}{!}{
  \begin{tabular}{lccccccccc}
    \toprule
    & \multicolumn{3}{c}{\textbf{Rubber}} &  \multicolumn{3}{c}{\textbf{PCB}} & \multicolumn{3}{c}{\textbf{Foam}} \\
    \cmidrule(lr){2-4}
    \cmidrule(lr){5-7}
    \cmidrule(lr){8-10}
    $F_{goal}$ & \multicolumn{1}{c}{$5N$} & \multicolumn{1}{c}{$10N$} & \multicolumn{1}{c}{$20N$} & \multicolumn{1}{c}{$1N$} & \multicolumn{1}{c}{$5N$} & \multicolumn{1}{c}{$8N$} & \multicolumn{1}{c}{$1N$} & \multicolumn{1}{c}{$1.5N$} & \multicolumn{1}{c}{$2N$} \\\midrule
    URScript &  \makecell{1.43\\-75\%} & \makecell{1.63\\-84\%} & \makecell{2.76\\-84\%} & \makecell{0.16\\-92\%} & \makecell{0.68\\-71\%} & \makecell{0.95\\-80\%} & \makecell{0.24\\-36\%} & \makecell{0.15\\-51\%} & \makecell{0.16\\-74\%} \\\midrule
    DMP &       \makecell{0.56\\-96\%} & \makecell{0.65\\-94\%} & \makecell{2.55\\-85\%} & \makecell{0.14\\-90\%} & \makecell{0.18\\-93\%} & \makecell{0.26\\-95\%} & \makecell{0.16\\-60\%} & \makecell{0.21\\-54\%} & \makecell{0.17\\-78\%} \\
    \bottomrule
   \end{tabular}}
   \caption{Experiment \ref{sec:experiment_2b}: Optimization of contact motions for different skill frameworks and surfaces. Each cell shows the mean deviation from $F_{goal}$ over 100 optimizations from random initial parameters (in $N$) and the improvement of this error over the initial parameterization.}
   \label{tb:mlrc_urscript_dmp_result}
\end{figure}

\subsection{Force-Sensitive Manipulation Without Expert Knowledge}
\label{sec:experiment_2}
\subsubsection{Data-driven optimization of contact forces} \label{sec:experiment_2a}
To evaluate our approach in the context of industrial manipulation, we consider the task of touching a surface with a specific impact force. In a first series of tests, we use \ac{spi} to optimize the motion direction, velocity $v$ and acceleration $a$ of a linear contact motion skill, which moves the robot in a given direction until a force $F_{goal}$ is registered. Contact motions are difficult to parameterize manually because the true contact force $F_{contact}$ is determined by a spring-mass-damper system composed of the robot and the contact surface with unknown dampening and spring characteristics. $F_{goal}$ merely imposes a lower bound on the maximum force, with $v$ and $a$ determining the true force on contact (cf. figure \ref{fig:results_mlrc} (gray)). A shadow model was trained on 50000 simulated and 500 real executions with randomly sampled values of $v$ and $a$. The task objective consisted of a linear combination of $\mathcal{G}_{cycle}$ and the mean squared error between the predicted contact force and $F_{goal}$. We ran \ac{spi} for goal forces between 3 and 7 $N$, collecting 250 inferred parameterizations for each goal force and randomly initializing $v$ and $a$. A total of 1250 optimized programs were executed on a Fanuc LR Mate 200iD/7L manipulator and FS-15iA \ac{fts}.\\
For goal forces between 3 and 5 $N$, the optimized parameterizations produce maximum contact forces very close to the target force (cf. figure \ref{fig:results_mlrc}, bottom left), demonstrating the capability of our approach to zero-shot generalize across task objectives (in this case, different values of $F_{goal}$). Figure \ref{fig:results_mlrc} (top right) illustrates the convergence behavior of our optimizer for a goal force of 5 $N$, which converges on a globally optimal velocity in under 40 \ac{nnii} iterations regardless of the initial parameterization. Figure \ref{fig:results_mlrc} (bottom right) shows the force trajectories resulting from executing the 250 resulting parameterizations.\\
\subsubsection{Generalization to different skill representations} \label{sec:experiment_2b}
In a second series of experiments, we use \ac{spi} over shadow skills to parameterize low-level primitives with respect to three different surfaces with very different dampening characteristics. To illustrate the universality of \ac{spi}, we optimize the target pose, velocity and acceleration parameters of the \texttt{movel} URScript primitive \cite{noauthor_urscript_2018} as well as the temporal scaling parameter $\tau$ and the target pose of a linear discrete \ac{dmp} \cite{ijspeert_dynamical_2013} to establish $F_{goal}$. The experiments were conducted on a Universal Robots UR5e. The results summarized in table \ref{tb:mlrc_urscript_dmp_result} show that the inferred parameterizations produce contact forces well within 0.25 $N$ of the goal on most surfaces, which is in the order of sensor noise. For both \acp{dmp} and URScript skills, \ac{spi} could adapt parameters to dampening characteristics ranging from near-linear (foam) to highly nonlinear (rubber) for a wide range of contact forces.

\subsection{Zero-Shot Generalization Across Task Objectives}
\label{sec:experiment_3}
\begin{figure}
  \centering
  \includegraphics[width=\linewidth]{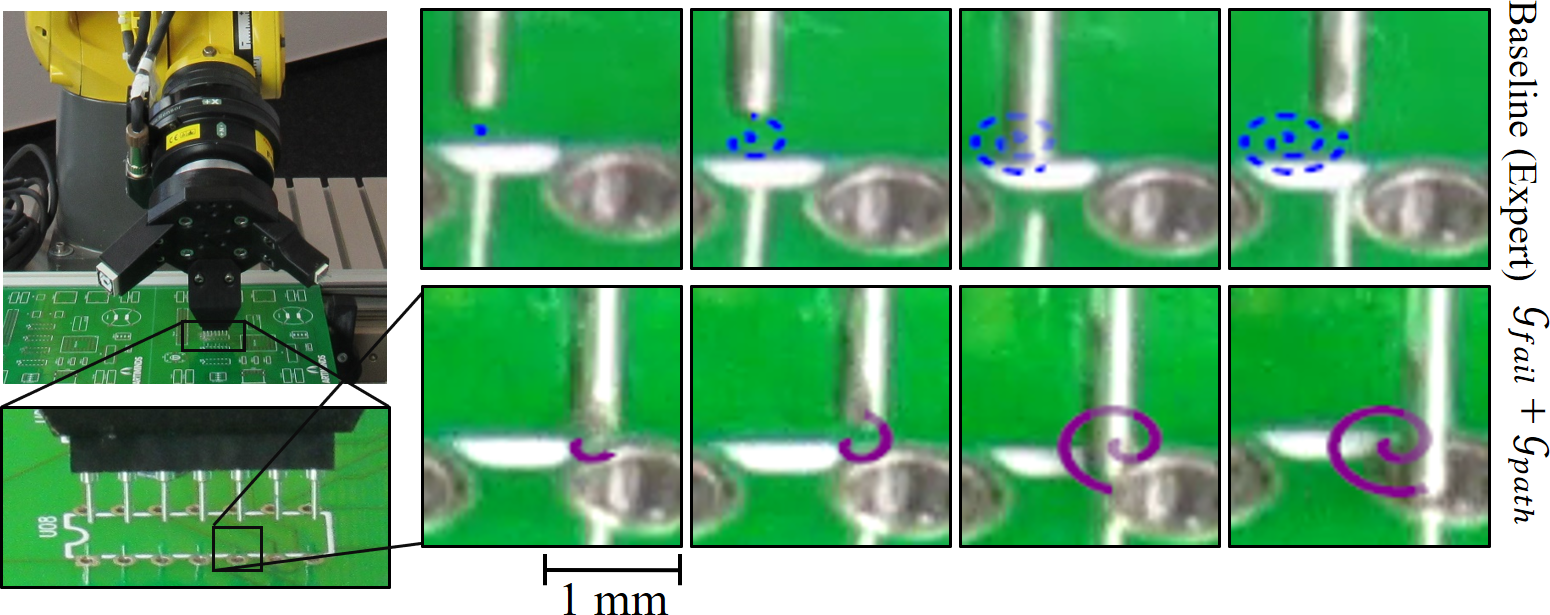}
  \vskip\baselineskip\vspace{-10pt}%
  \includegraphics[width=\linewidth]{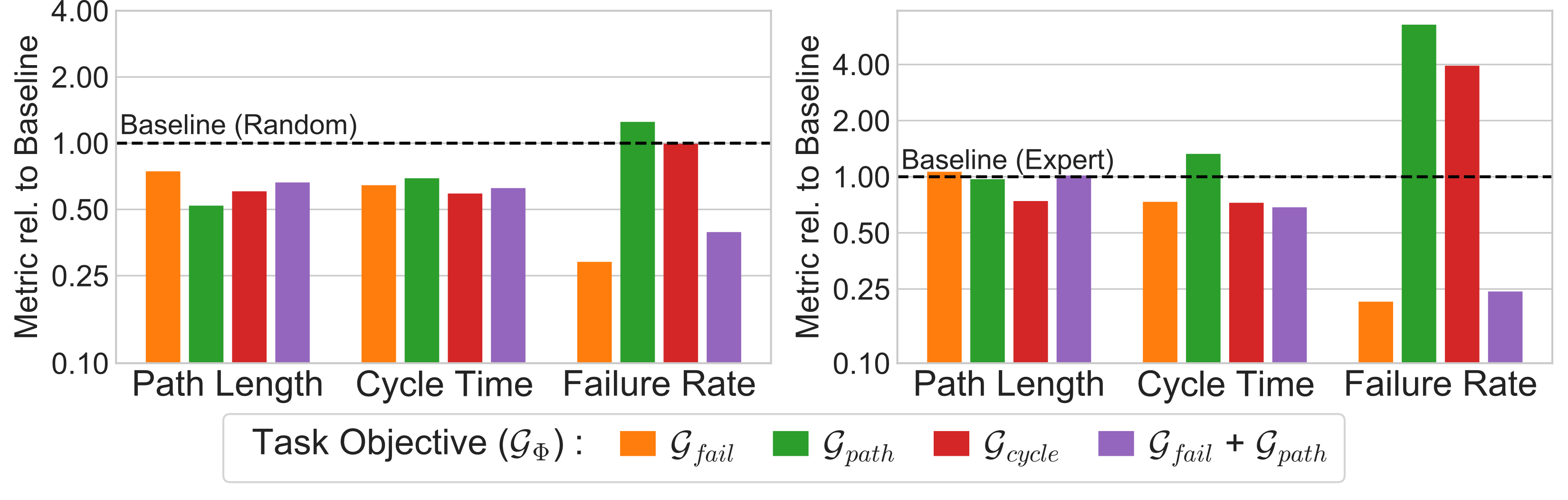}  \vskip\baselineskip\vspace{-10pt}%
  \includegraphics[width=\linewidth]{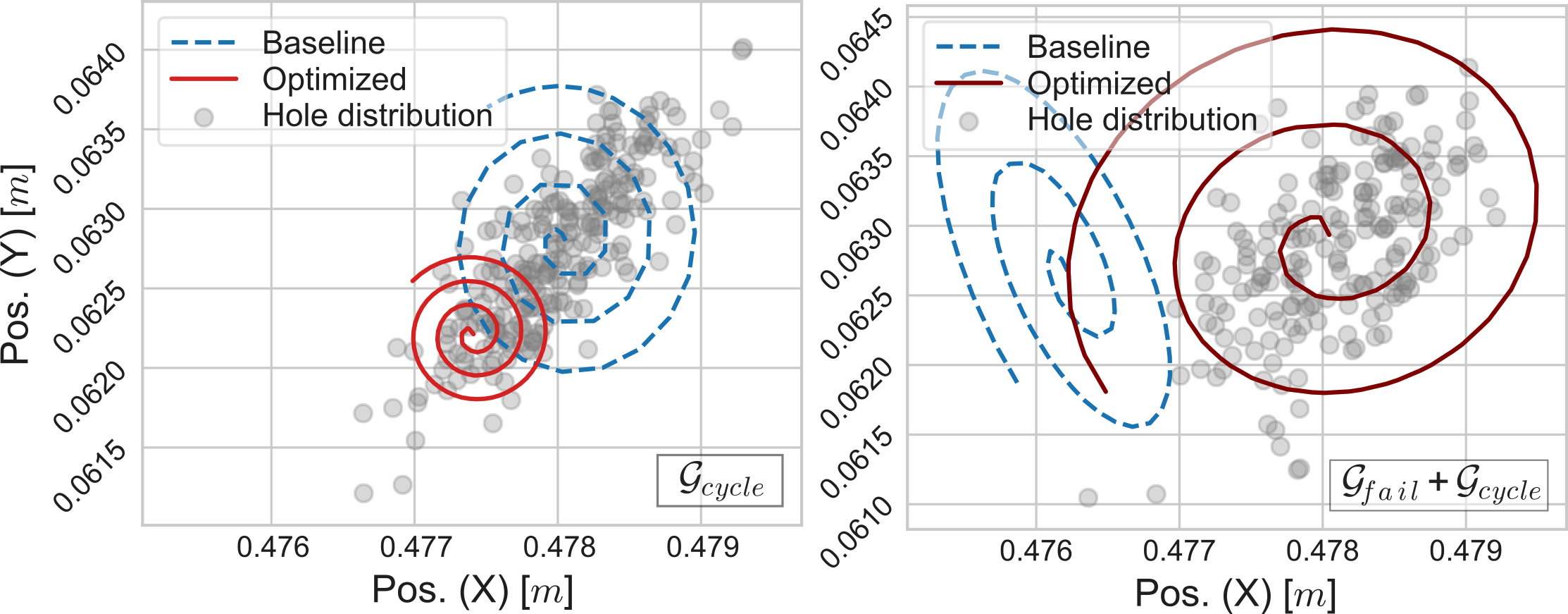}
  \caption{Experiment \ref{sec:experiment_3}: Top: Stochastic variations of the hole position cause a manually parameterized spiral search to fail, while an inferred parameterization is more robust. Middle: Process metrics for optimized programs relative to random (left) and expert (right) baselines (log scale). Bottom: Spiral search policies for different objective functions.}
  \label{fig:results_spiral}
\end{figure}

To illustrate the capacity of \ac{spi} to zero-shot generalize across task objectives, we consider a further use case of finding the position of a set of holes on a \ac{pcb} for the insertion of electronics components. In practice, manufacturing tolerances cause stochastic deviations from the expected hole positions on the order of millimeters, requiring the use of force-controlled search motions. An program structure to solve this task consists of a linear approach motion followed by a force-controlled spiral search (cf. figure \ref{fig:neural_program}). The spiral search skill accepts inputs $w_x$ and $w_y$ defining the extents of the spiral motion along its principal axes, the distance $d$ between spiral arms, force runtime constraints ${F_{min}, F_{max}}$, position goal constraints ${z_{min}, z_{max}}$, velocity $v$ and acceleration $a$ and performs a spiral motion in the xy-plane of the \ac{tcp}, maintaining a force between $F_{min}$ and $F_{max}$ along the z-axis of the \ac{tcp}, succeeding if a depth between $z_{min}$ and $z_{max}$ can be reached. Shadow models for both skills were pre-trained on 50000 simulated executions and fine-tuned on 2500 real samples using a Fanuc LR Mate 200iD/7L and FS-15iA \ac{fts}. We collected two baseline test datasets of 250 samples each, one with randomly initialized input parameters and one parameterized by a human expert. Parameter inference was conducted from initial input parameters set to the respective baseline parameterization. To demonstrate zero-shot generalization across task objectives, program parameters were optimized with respect to $\mathcal{G}_{cycle}$, $\mathcal{G}_{fail}$, $\mathcal{G}_{path}$ (cf. \ref{eq:g_cycle}) as well as linear combinations.\\
The optimized parameterizations consistently yield improvements for their corresponding metrics (cf. figure \ref{fig:results_spiral}). Compared to the already robust human expert parameterization, the optimized parameterization nearly eliminates failures altogether. Joint parameter inference with respect to a combination of task objectives yields gains in both metrics. Figure \ref{fig:results_spiral} shows examples for spiral motions resulting from the optimized policies. Optimization with respect to different task objectives results in fundamentally different search policies, such as a ``fail fast'' policy for minimizing path length or a  very robust policy for minimizing failure rate and cycle time which near-optimally fits the hole distribution.

\section{Conclusion and Outlook}
\label{sec:conclusion}
We present an approach for inferring the input parameters of skill-based robot programs by a combination of unsupervised learning and gradient-based iterative model inversion. To our knowledge, this is the first application of \acl{dp} and \ac{nnii} to the skill parameterization problem in robotics. We show that \ac{spi} can effectively infer optimal parameters for robot programs composed from  non-differentiable skill frameworks. Application to force-sensitive contact motions and search heuristics for electronics assembly demonstrate its capability to adapt parameters to nonlinear system dynamics or stochastic process noise. Zero-shot generalization across task objectives and exclusive reliance on unsupervised training establish \ac{spi} as a powerful solution for parameter inference in real-world use cases.\\
Like all first-order optimization methods, the performance of \ac{spi} is conditional on the topology of the objective function. We are investigating the possibility of augmenting \ac{nnii} by hessian-free optimization to further improve performance \cite{martens_deep_2010}. We further observe that \ac{rl}-based robot learning approaches require a solution to locally optimize skill parameters \cite{englert_learning_2018, englert_combined_2016}, motivating further inquiry on synergies between \ac{spi} and \ac{rl} for robot 
learning.
\section*{Acknowledgment}
This work was supported by the Federal Ministry of  Education  and  Research  (BMBF)  under  grant  no. 01DR19001B.






\bibliographystyle{IEEEtran_bst/IEEEtran}
\bibliography{icra_2021_alt}

\end{document}